\documentclass[10pt,twocolumn,letterpaper]{article}

\usepackage{icb}
\usepackage{times}
\usepackage{epsfig}
\usepackage{graphicx}
\usepackage{amsmath}
\usepackage{amssymb}
\usepackage{xcolor}  % TODO: remove this line before submitting
\usepackage{multirow}
\usepackage{enumerate}
\usepackage{url}
\usepackage{graphicx}
\usepackage{caption}
\usepackage{subcaption}
\usepackage{tabularx}
\usepackage{enumitem}
\usepackage[norule,symbol,perpage]{footmisc}

\makeatletter 
\def\ps@IEEEtitlepagestyle{ 
\def\@oddfoot{\mycopyrightnotice} 
\def\@evenfoot{} 
} 
\def\mycopyrightnotice{ 
{\hfill \footnotesize 978-1-7281-3640-0/19/\$31.00 \copyright 2019 IEEE\hfill} 
} 
\makeatother

\usepackage[pagebackref=true,breaklinks=true,letterpaper=true,colorlinks,bookmarks=false]{hyperref}

\icbfinalcopy % *** Uncomment this line for the final submission

\ificbfinal\fi
\begin{document}

%%%%%%%%% TITLE
\title{Generalized Presentation Attack Detection: \\a face anti-spoofing evaluation proposal}

\author{Artur Costa-Pazo\textsuperscript{\textdagger}\textsuperscript{\textsection}, David Jim\'enez-Cabello\textsuperscript{\textdagger}, Esteban Vazquez-Fernandez\textsuperscript{\textdagger}\\
GRADIANT, Vigo (Spain)\textsuperscript{\textdagger}\\
{\tt\small \{acosta,djcabello,evazquez\}@gradiant.org }
\and
% David Jim\'enez-Cabello\textsuperscript{\textdagger}\\
% GRADIANT, Vigo (Spain)\textsuperscript{\textdagger}\\
% {\tt\small djcabello@gradiant.org }
% \and
% Esteban Vazquez-Fenandez\textsuperscript{\textdagger}\\
% GRADIANT, Vigo (Spain)\textsuperscript{\textdagger}\\
% {\tt\small evazquez@gradiant.org }
% \and
Jos\'e Luis Alba-Castro\textsuperscript{\textsection}\\
University of Vigo (Spain)\textsuperscript{\textsection}\\
{\tt\small jalba@gts.uvigo.es }
\and
Roberto J. L\'opez-Sastre\textsuperscript{\textdaggerdbl}\\
University of Alcal\'a (Spain)\textsuperscript{\textdaggerdbl}\\
{\tt\small robertoj.lopez@uah.es }
}

% \author{Artur Costa-Pazo$^\textsuperscript{1 2}$, David Jim\'enez-Cabello$^\textsuperscript{1}$, Esteban Vazquez-Fenandez$^\textsuperscript{1}$,\\ Jos\'e Luis Alba-Castro$^\textsuperscript{2}$ and Roberto J. L\'opez-Sastre$^\textsuperscript{3}$ \\
% GRADIANT, Vigo (Spain)$^\textsuperscript{1}$, University of Vigo (Spain)$^\textsuperscript{2}$, University of Alcal\'a (Spain)$^\textsuperscript{3}$\\
% {\tt\small \{acosta,djcabello,evazquez\}@gradiant.org, jalba@gts.uvigo.es, robertoj.lopez@uah.es}
% }

\maketitle
\thispagestyle{empty}

\begin{abstract}
 % Over the past few years, Presentation Attack Detection (PAD) has become a fundamental part of facial recognition systems. Although much effort has been devoted to the investigation of anti-spoofing techniques, the main challenge is still their generalization capability, what we call Generalized Presentation Attack Detection (GPAD). In this paper we present a new open-source evaluation framework to advance in the investigation of face-GPAD. It allows to conduct experiments and facilitate the creation of new protocols focused on the generalization problem, to establish fair procedures of evaluation and comparison between PAD solutions. We also introduce the largest aggregated and categorized data set, to address the problem of incompatibility between publicly available datasets. Finally, we propose a clear benchmark, to help researchers to deal with the  generalization issues, proposing two novel evaluation protocols: the first is oriented to measure the effect introduced by the variations in face resolution, and the second consists in evaluating the influence of adversarial operating conditions.

 Over the past few years, Presentation Attack Detection (PAD) has become a fundamental part of facial recognition systems. Although much effort has been devoted to anti-spoofing research, generalization in real scenarios remains a challenge. In this paper we present a new open-source evaluation framework to study the generalization capacity of face PAD methods, coined here as face-GPAD. This framework facilitates the creation of new protocols focused on the generalization problem establishing fair procedures of evaluation and comparison between PAD solutions. We also introduce a large aggregated and categorized dataset to address the problem of incompatibility between publicly available datasets. Finally, we propose a benchmark adding two novel evaluation protocols: one for measuring the effect introduced by the variations in face resolution, and the second for evaluating the influence of adversarial operating conditions.
\end{abstract}

\section{Introduction}

Face recognition based systems have become very attractive solutions for a wide variety of applications. The presentation of the iPhone X in 2017 with its face verification system (the “FaceID”) puts this technology in the spotlight as a strong candidate to substitute the fingerprint verification, not only to unlock the device but also as a mobile authentication mechanism.% to prevent spoofing attacks.
 Furthermore, these systems are increasingly used in several applications as border controls, accesses to events or even for on-boarding processes. %These systems are highly valued by consumers because of their usability and its non-intrusive nature. 
However, there remain two major challenges for the inclusion of these kind of systems in a larger number of applications: the Presentation Attacks (PAs) and their generalization capability. 

% PAs are still the greatest vulnerability of face recognition systems. Nowadays, it is really easy to spoof most of the face recognition systems using social networks as a source to get target user images and videos. The most common attacks that have been studied in the literature are those based on the following two types of Presentation Attack Instruments (PAIs): i) those based on printed material (the image of the target user is presented on printed paper); or ii) those based on replay (image or video is presented on a screen). Moreover, in recent years the emergence of different technologies (e.g. 3D face reconstruction, custom-made flexible silicone masks, realistic and dynamic face avatars, etc.) has made it easier and easier to create much more sophisticated attacks.

%Therefore, t
The development of Presentation Attack Detection (PAD) solutions is a must, to guarantee the security of the users. Early adopters of face recognition (FR) systems have historically chosen \textit{active face-PAD} approaches in order to preserve their security. This challenge-response strategy requires to mimic certain actions (e.g. eye blinking, smiling, head motion) increasing its robustness to attacks based on photographs or videos. However, latest advances in computer graphics \cite{jackson2017vrn, Korshunova_2017_ICCV} threaten even this collaborative countermeasure. On the other hand, \textit{passive face-PAD} is a more convenient and non-intrusive approach, but also entailing a far more challenging scenario. Relying on additional sensors such as 3D/IR/thermal cameras is an easy option, but it restricts the use case to a few specialized devices, dramatically incrementing costs. For the sake of accessibility and costs, we focus on the ubiquitous 2D-camera case, available in almost all mobile devices. %and easy to acquire and integrate on different checkpoints.

On the other hand, we find the generalization problem. After a first stage of enthusiasm among the research community, with PAD solutions that offer very low error rates in the available datasets, it comes a stage of disappointment when the same approaches are not able to work in more real environments. For instance, cross-evaluations between datasets quickly began to be included in the publications \cite{can_face_antispoofing_defreitas2013} in order to reveal a great problem: methods evaluated in a dataset other than training increase error rates by an order of magnitude. However, there is no common evaluation framework for conducting such cross-evaluation experiments in a consistent and comparable manner across publications. Therefore, addressing the generalization problem is critical.

We believe that generalization in face-PAD is in an early research stage and it is time to unify the criteria in order to perform fair evaluations. Thus, in this paper we propose to establish a new topic within the anti-spoofing research that deals with the generalization issues within face-PAD methods. We call this problem Generalized Presentation Attack Detection (GPAD).

We propose to analyze the status quo of the face anti-spoofing research with a proposal centered around the GPAD problem. The main contributions of this work include:
\begin{itemize}[leftmargin=7mm]
\setlength{\itemsep}{-0.27\baselineskip}
\item We provide the largest aggregated dataset with a common categorization in two levels to represent four key aspects in anti-spoofing: attacks, lighting, capture devices and resolution.
\item We release an open-source evaluation framework\footnote{https://github.com/Gradiant/bob.paper.icb2019.gradgpad}, introducing an unified benchmark for GPAD. 
\item We provide an evaluation of state-of-the-art methods in the proposed benchmark. We demonstrate the limitation of current dataset evaluation procedures (generalization, cross-domain evaluation, etc.), while showing the benefits of the proposed unified framework. All the experiments will be reproducible.
\item Using the novel evaluation tool, we introduce two novel protocols for the GPAD problem.%: the first is used to measure the effect introduced by the variations in face resolution, and the second focuses on the evaluation of the influence of adversarial operating conditions.
\end{itemize}

The remainder of this paper is organized as follows. Section~\ref{sec:related} provides a brief overview of the challenges in the field of generalization for anti-spoofing systems. The proposed framework and the description of the  Aggregated-Datasets are presented in Section~\ref{sec:proposal}. In Section~\ref{sec:experiments} we discuss in detail the proposed evaluation protocols and results. Conclusions are presented in Section~\ref{sec:conclusions}. 

%----------------------------------------------------------
\section{Related Work}
\label{sec:related}

As \cite{boulkenafet2018generalization} already points out, there is no unified taxonomy for the diverse set of attacks, nor for a unified criteria for the evaluation protocols or the metrics used to fairly assess the performance of face-PAD systems, specially when dealing with the generalization problem. We compliment the taxonomy initiated in \cite{boulkenafet2018generalization} to fully categorize the current status of face-PAD approaches from different perspectives.

Current face-PAD methods can be classified regarding the following standpoints: i) from the hardware used for data acquisition as \textit{rgb-only} \cite{iqa_galbally_2015, anomaly_idiap_2018, db_msu_mfsd_2015} or \textit{additional sensors} \cite{db_csmad_2018, light_field_based_sepas_2018} approaches; ii) from the required user interaction as \textit{active} \cite{kollreider2007real} or \textit{passive} \cite{li2018learning, db_msu_mfsd_2015} methods; iii) from the input data type as \textit{single-frame} \cite{db_msu_mfsd_2015} or \textit{video-based} \cite{motion_anjos_2011, db_uvad_2015} approaches; iv) and, finally, depending on the feature extraction and classification strategy as \textit{hand-crafted} \cite{boulkenafet2018generalization, db_msu_mfsd_2015} or \textit{deep learning} \cite{db_siw_2018, li2018learning}. Based on these classifications, we can depict that the most challenging scenario occurs when data is captured using rgb-only sensors %, incorporated in almost every mobile device for instance, and 
using passive approaches that avoid any challenge-response interaction with the user (e.g. smiling or blinking) that increases drastically the usability of face-PAD systems. The most recent and successful methods \cite{db_siw_2018, db_rose_youtu_2018, wang2018exploiting} show that video-based solutions incorporate more meaningful information compared with those based on single frames, adopting the former as the main research direction. Finally, despite that many recent models attempt to detect face liveness, building upon representation based hand-crafted features \cite{iqa_galbally_2015, maatta2011face, tirunagari2015detection}, obtaining good results for intra-dataset protocols, the increasing interest on the topic has led to the appearance of new datasets \cite{db_siw_2018, db_rose_youtu_2018, db_casia_surf_2018}, turning deep learning methods as a solid alternative adopted by almost every recent approach \cite{ db_siw_2018, li2018learning, db_rose_youtu_2018, db_smad_2017, db_casia_surf_2018}.
Considering the aforementioned challenging setting (rgb-only, passive, video-based and deep learning), these methods leverage recent advances on deep metric learning to extract highly discriminative features from image sequences, achieving state-of-the-art results for intra-dataset setups, especially when using auxiliary supervision based on depth reconstruction \cite{db_siw_2018, li2018learning}. Despite training data is not enough, this setting has been established as the right direction to discover the complex patterns and details that appear in any spoofing attempt.

%de2012lbp

\subsection{Generalized Presentation Attack Detection}

Current state-of-the-art solutions suffer a severe drop of performance during testing in realistic scenarios %to assess generalization capabilities,
because they exhibit a sort of overfitting behavior maximizing the results for intra-dataset experimental setups. This fact has made that each approach proposes its own evaluation protocols and metrics, giving place to a missing unified criteria to evaluate generalization.

Generalization has been specifically addressed with no success from different perspectives: i) applying \textit{domain adaptation} techniques \cite{db_rose_youtu_2018}; ii) reformulating the problem as an \textit{anomaly detection} \cite{anomaly_idiap_2018} scenario; iii) learning \textit{generalized deep features} \cite{db_siw_2018, li2018learning, db_rose_youtu_2018}; or even iv) using generative models \cite{db_siw_2018}. Besides, there is still a lack of unified benchmark and representative datasets, that might mitigate the constant improvement of Presentation Attack Instruments (PAIs) with more sophisticated strategies (e.g. 3D rendering) and newly unseen attack instruments (e.g. professional make-up). Regardless almost every proposal comes with its own reduced dataset \cite{db_siw_2018, db_rose_youtu_2018, db_smad_2017, db_casia_surf_2018}, there is \textit{no agreement upon a PAD benchmark}, and generalization properties are not properly evaluated. 

\subsection{Datasets}

During a brief inspection of the capture settings of video-based face-PAD datasets (see Table \ref{table:datasets}), one can easily observe that there is no unified criteria in the goals of each of them, leading to a manifest built-in bias. This specificity in the domain covered by most of the datasets can be observed in different scenarios: i) some of them focus on a single type of attack (e.g masks - 3DMAD, HKBU, CSMAD), ii) others focus  on the study of different image sources (depth/NIR/thermal) such as CASIA-SURF or CSMAD, iii) others attempt to simulate a certain scenario like a mobile device setting, where the user hold the device (e.g. Replay-Mobile, OULU-NPU), or a webcam setting, where user is placed in front of fixed camera (e.g Replay-Attack, SiW), or even a stand-up scenario where users are recorded further from the camera (e.g UVAD), etc. 

\begin{table*}[]
\tiny
\centering
\begin{tabular}{|c|c|c|c|c|c|c|c|c|c|c|}
\hline
Dataset         & Year & \begin{tabular}[c]{@{}c@{}}Num \\ Identities\end{tabular} & \begin{tabular}[c]{@{}c@{}}Number samples\\ real/ attack\end{tabular} & \begin{tabular}[c]{@{}c@{}}Spoof\\ attacks\end{tabular}                       & \begin{tabular}[c]{@{}c@{}}Capture\\ Devices\end{tabular}                                                                                                                         & \begin{tabular}[c]{@{}c@{}}Modal\\ types\end{tabular} & \begin{tabular}[c]{@{}c@{}}Display\\ Devices\end{tabular}                                  & \begin{tabular}[c]{@{}c@{}}Pose\\ Range\end{tabular}                & \begin{tabular}[c]{@{}c@{}}Different\\ Expression\end{tabular} & \begin{tabular}[c]{@{}c@{}}Additional\\ Lightning\end{tabular} \\ \hline \hline
CASIA-FASD~\cite{db_casia_fasd_2012}     & 2012 & 50                                                        & 150/450                                                               & \begin{tabular}[c]{@{}c@{}}Print, \\ Replay\end{tabular}                      & \begin{tabular}[c]{@{}c@{}}low-quality webcam,\\  medium-quality webcam, Sony NEX-5\end{tabular}                                                                               & RGB                                                   & iPad                                                                                       & Frontal                                                             & No                                                             & No                                                             \\ \hline
REPLAY-ATTACK~\cite{db_replay_attack_2012}   & 2012 & 50                                                        & 200/1000                                                              & Print, 2 Replay                    & macbook webcam                                                                                                                                                                    & RGB                                                   & iPhone 3GS, iPad                                 & Frontal                                                             & No                                                             & Yes                                                            \\ \hline
3DMAD~\cite{db_3dmad_2013}            & 2013 & 17                                                        & 170/85                                                                & Mask (rigid)                                                                          & Microsoft Kinect                                                                                                                                                                  & RGB/Depth                                             & -                                                                                          & Frontal                                                             & No                                                             & No                                                             \\ \hline
MSU-MFSD~\cite{db_msu_mfsd_2015}        & 2015 & 35                                                        & 110/330                                                               & Print, 2 Replay                     & macbook air webcam, nexus 5                                                                                                              & RGB                                                   & iPad Air, iPhone 5S                              & Frontal                                                             & No                                                             & No                                                             \\ \hline
UVAD~\cite{db_uvad_2015}           & 2015 & 404                                                       & 808/16268                                                             & 7 Replay                                                                      & \begin{tabular}[c]{@{}c@{}}cybershot\_dsc-hx1, canon\_powershot\_sx1, \\ nikon\_coolpix\_p100, kodak\_z981, \\ olympus\_sp\_800UZ, panasonic\_fz35\_digital\end{tabular} & RGB                                                   & \begin{tabular}[c]{@{}c@{}}7 different \\ unknown monitors\end{tabular}                    & Frontal                                                             & No                                                             & No                                                             \\ \hline
REPLAY-MOBILE~\cite{db_replay_mobile_2016}   & 2016 & 40                                                        & 390/640                                                               & Print, Replay                       & iPad Mini 2, LG G4                                                                                                                     & RGB                                                   & Philips 227ELH                                                                             & Frontal                                                             & No                                                             & Yes                                                            \\ \hline
HKBU~\cite{db_hkbu_2016} (v1)       & 2016 & 8                                                         & 70/40                                                                 & Mask (rigid)                                                                          & Logitech C9200 webcam                                                                                                                                                             & RGB                                                   & -                                                                                          & Frontal                                                             & No                                                             & No                                                             \\ \hline
OULU-NPU~\cite{db_oulu_npu_2017}        & 2017 & 55                                                        & 1980/3960                                                             & \begin{tabular}[c]{@{}c@{}}2 Print,\\ 2 Replay\end{tabular}                   & \begin{tabular}[c]{@{}c@{}}Samsung Galaxy S6 Edge, htc\_desire\_eye, \\ meizu\_x5, asus\_xenfone, sony\_xperia\_c5, oppo\_n3\end{tabular}                             & RGB                                                   & \begin{tabular}[c]{@{}c@{}}Dell1905FP,\\ MacBook Retina\end{tabular}                       & Frontal                                                             & No                                                             & Yes                                                            \\ \hline
SMAD~\cite{db_smad_2017}           & 2017 & -                                                         & 65/65                                                                 & Mask (silicone)                     & -                                                                                                                                                                                 & RGB                                                   & -                                                                                          & -                                                                   & -                                                              & -                                                              \\ \hline
ROSE-YOUTU~\cite{db_rose_youtu_2018}      & 2018 & 20                                                        & 3350                                                                  & \begin{tabular}[c]{@{}c@{}}2 Print, 2 Replay,\\ 2 Mask (Paper)\end{tabular} & \begin{tabular}[c]{@{}c@{}}hasee, huawei, ipad\_4, iphone\_5s, zte\end{tabular}                                                                                       & RGB                                                   & \begin{tabular}[c]{@{}c@{}}Lenovo LCD display, \\ Mac LCD display\end{tabular}             & Frontal                                                             & Yes                                                            & No                                                             \\ \hline
SIW~\cite{db_siw_2018}            & 2018 & 165                                                       & 1320/330                                                              & \begin{tabular}[c]{@{}c@{}}2 Print,\\ 4 Replay\end{tabular}                   & \begin{tabular}[c]{@{}c@{}}canon\_eos\_t6, logitech\_c920\_webcam\end{tabular}                                                                                                 & RGB                                                   & \begin{tabular}[c]{@{}c@{}}iPad Pro, iPhone7\\ Galaxy S8, Asus MB168B\end{tabular} & {[}-90º, 90º{]}                                                     & Yes                                                            & Yes                                                            \\ \hline
CS-MAD~\cite{db_csmad_2018}          & 2018 & 14                                                        & 88/220                                                                & \begin{tabular}[c]{@{}c@{}}Print,Mask \\ (silicone)\end{tabular}              & \begin{tabular}[c]{@{}c@{}}Intel Realsense SR300 (full HD), \\ Nikon Coolpix P520 (still images)\end{tabular}                                                                 & \begin{tabular}[c]{@{}c@{}}RGB/Depth \\ IR/Thermal   \end{tabular}                               & -                                                                                          & Frontal                                                             & No                                                             & Yes                                                            \\ \hline
CASIA-SURF~\cite{db_casia_surf_2018}     & 2019 & 1000                                                      & 3500/17500                                                            & \begin{tabular}[c]{@{}c@{}}Print, Mask \\ (paper, cut)\end{tabular}         & Intel RealSense                                                                                                                                                                   & \begin{tabular}[c]{@{}c@{}} RGB/Depth \\IR  \end{tabular}                                        & -                                                                                          & \begin{tabular}[c]{@{}c@{}}{[}-30º, 30{]}\\(attacks)\end{tabular} & No                                                             & No                                                             \\ \hline
\end{tabular}
\caption{List of existing databases for anti-spoofing based on videos and their main characteristics.}
\label{table:datasets}
\end{table*}

Beyond classical intra-dataset evaluation where state-of-the-art algorithms work really well, most of them also propose to assess generalization properties using a direct evaluation between different datasets (i.e. inter-dataset protocols). The most recent publications go a step further by proposing aggregated datasets such as \cite{anomaly_idiap_2018} and \cite{face-upad}. The underlying idea is to generate a combined dataset that represents the anti-spoofing scenario in a richer way, increasing the diversity of the data, e.g. with a larger number of users, capture devices and PAIs. 

It seems clear that the community has become aware of the importance of data representativeness in our topic and more and more publications are proposing new databases or different combinations of inter-dataset evaluations. %However, certain evaluations are obsolete and it is necessary to reformulate the way in which we capture and evaluate our data.

\subsection{Challenges}

Despite the effort of the research community, there is no standard methodology to evaluate the different approaches in a cross-dataset scenario, which motivates each publication to propose a new method with a different dataset. This makes it impossible to perform a more in-depth analysis of anti-spoofing algorithms, since by keeping the global results of a dataset, we are masking many variables. Thus, some questions quickly arise. Are our algorithms working badly for an specific type of attack, capture device or resolution?, or is it really just one of those hidden parameters that is causing that the performance drops drastically? It is obvious that there is a big challenge under the generalization problem and we need to unmask it in detail if we aim at giving a big step towards a general solution. We propose to emphasize the importance of data and to unify the procedure to evaluate face anti-spoofing systems in the most challenging scenario: the generalization problem. %To this end, we propose, to the best of our knowledge, not only the largest aggregated dataset at the moment, but also an evaluation framework that allows a fair comparison between the different approaches provided by a general categorization of the attributes that are shared among most of the existing datasets.

%----------------------------------------------------------
\section{The Evaluation Framework}
\label{sec:proposal}

%\subsection{Purpose}

In this paper, we present a novel evaluation framework to help the research community to address the problem of generalization in face-PAD. Our framework is publicly available as a Python package under the acronym GRAD-GPAD (Generalization Representation over Aggregated Datasets for Generalized Presentation Attack Detection). This framework provides a common taxonomy of existing datasets and allows to evaluate face-PAD algorithms from additional points of view, revealing new interesting properties. The GRAD-GPAD framework is extendable and scalable by design. This will help researchers to create new protocols, to define categorizations over different datasets in order to make them compatible, and to enable the addition of new datasets.

\begin{figure}
\includegraphics[width=\linewidth]{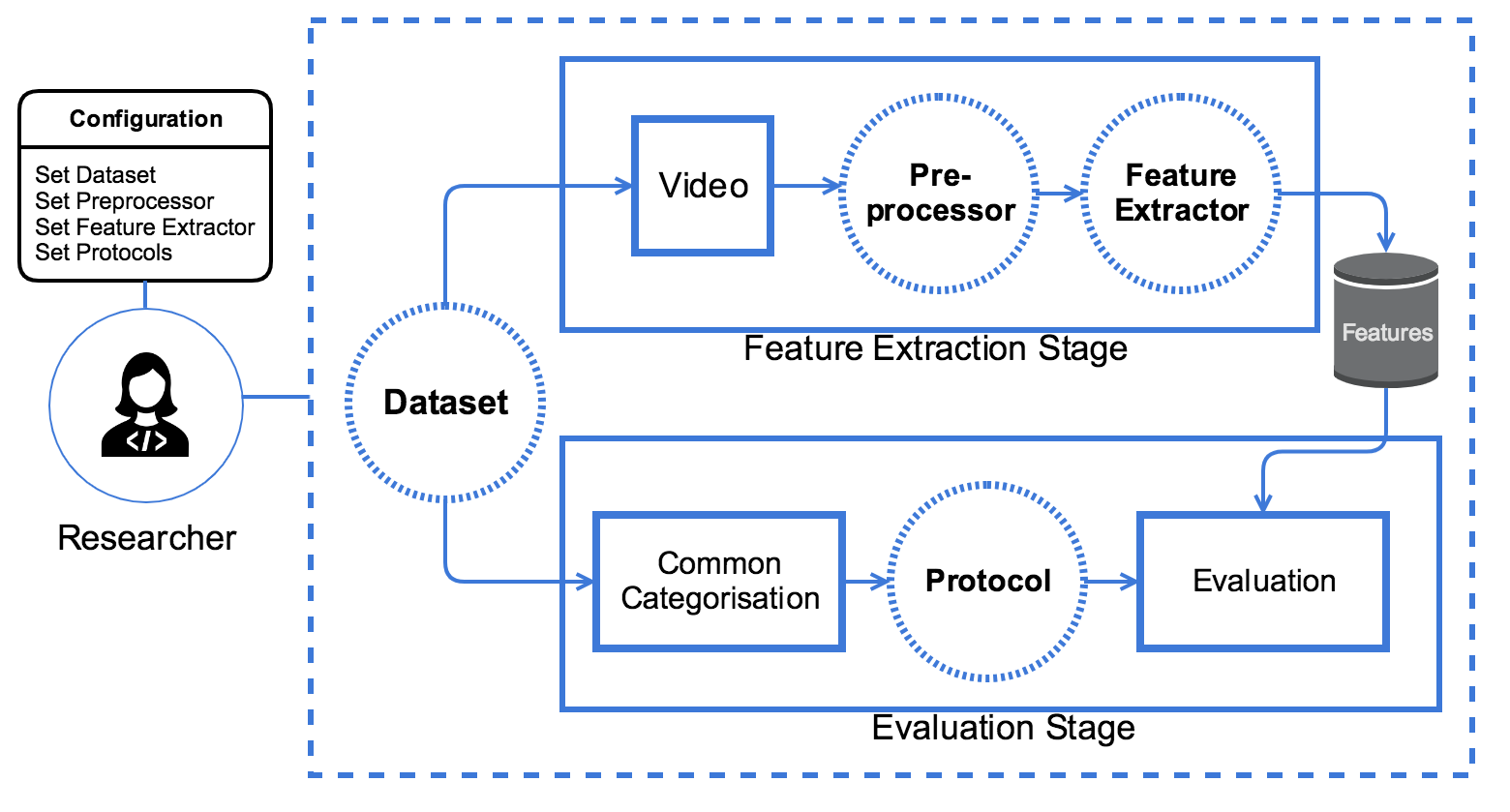}
\caption{Overview of GRAD-GPAD evaluation framework}
\label{figure:frameworkdiagram}
\end{figure}

The GRAD-GPAD framework has a top-down design and provides a very convenient level of abstraction. Figure \ref{figure:frameworkdiagram} shows the proposed dataset evaluation pipeline, where circles with dotted lines represent the customizable and easy-to-extend elements. Each of these elements in the framework follows a specific interface to work properly. Thus, we can add, for instance, a new \textit{Dataset} (e.g. using some videos recorded for an specific use case), or a new implementation of a \textit{Feature Extractor} to evaluate an algorithm or even create an ad-hoc \textit{Protocol}. 

GRAD-GPAD presents two main stages: i) \textit{feature extraction}, where features are computed for each input video applying a preprocessing step and following the interface of the Feature Extractor; and ii) \textit{evaluation}, where a filtering step is applied, using the already extracted features and the common categorization of the datasets, to train and test over the selected features.

\subsection{The Aggregated Dataset}

We propose a novel paradigm of data collection where new datasets are aggregated in a larger  dataset, following a global categorization to make the involved datasets compatible. We show that our framework can be used to replicate and extend the analysis of state-of-the-art protocols, as well as to obtain a fair analysis of the algorithms using novel protocols.

As we already pointed out in Section 2, current publicly available datasets for face anti-spoofing are very heterogeneous. The variability of the spoofing attacks and capture/display devices makes the problem actually unconstrained and dynamic. Moreover, existing datasets present different labeling, protocols and data splits. The GRAD-GPAD framework simplifies such a dynamic structure thanks to its scalable nature and only requires the data to be split into three subsets in order to evaluate face-PAD algorithms: a training set to find the classification model, the development set to estimate the threshold that provides Equal Error Rate (EER) performance, and a  test set that is used to report the final results using fair metrics for the generalization problem that has been recently standardized in the \textit{ISO/IEC 30107-3} \footnote{https://www.iso.org/standard/67381.html}: i.e. HTER (\textit{Half Total Error Rate}), ACER (\textit{Average Classification Error Rate}), APCER (\textit{Attack Presentation Classification Error Rate}) and BPCER (\textit{Bona fide Presentation Classification Error Rate}). Some datasets (e.g. Replay-Attack \cite{db_replay_attack_2012}, 3DMAD \cite{db_3dmad_2013}, Replay-Mobile \cite{db_replay_mobile_2016}, HKBU \cite{db_hkbu_2016}, OULU-NPU \cite{db_oulu_npu_2017}) are already divided that way, others, however, need to be restructured. Based on the work of Pinto et al. in \cite{db_uvad_2015}, we have split CASIA-FASD \cite{db_casia_fasd_2012}, Rose-Youtu \cite{db_rose_youtu_2018} and SiW \cite{db_siw_2018}, keeping the test subset unmodified and splitting the original training set in a training subset comprising 80\% of the users and a development subset comprising the remaining 20\%. Furthermore, CS-MAD \cite{db_csmad_2018} does not contain explicit subsets, so we randomly partitioned the data into the mentioned subsets (40\% in Train, 30\% in Dev and 30\% in Test) from the users' identities. Finally MSU-MFSD \cite{db_msu_mfsd_2015} is originally divided in two folds, nevertheless we re-divided it based on this Python package\footnote{https://gitlab.idiap.ch/bob/bob.db.msu\_mfsd\_mod}. At the moment of writing this paper, ten out of the thirteen datasets shown in Table \ref{table:datasets} have been integrated in our framework. Work is currently underway to add both SMAD \cite{db_3dmad_2013} and UVAD \cite{db_uvad_2015}, as well as the recently introduced CASIA-SURF \cite{zhang2018casia}. 

\subsection{Categorization}

Although some of the datasets share a common notation, there is no common taxonomy,  so we can categorize the different types of accesses represented in current datasets.  Each of these datasets has useful categories and labels that help researchers for a better understanding of their algorithms. However, the amount of data is usually not enough to train the algorithms properly, so they cannot be effectively used. In this paper, we propose an updated inter-dataset categorization based on four general categories: \textit{common PAI}, \textit{common capture device}, \textit{common lighting} and \textit{common face resolution}.
% i) \textit{common PAI}, that represents the attacks available on the aggregated dataset (print, replay and mask); ii) \textit{common capture device}, that represents the variability of capture devices splitted into three categories (webcam, mobile/tablet and digital camera); iii) \textit{common lighting}, based on illumination information already labeled as controlled and adverse conditions; and iv) \textit{common face resolution}, that categorizes the access based on the size of the inter eye distance of the detected faces, subdividing them into three types (small, medium and large face). 
Table \ref{table:categorization} and the next paragraphs present, in more detail, each of the proposed categories, as well as their subclasses (type and subtype).

\begin{table}[htbp]
\tiny
\centering
\begin{tabular}{llll}
Category                                                      & Types                                                & Sub-type                       & Criteria                                                  \\ \hline
\multicolumn{1}{|l|}{\multirow{9}{*}{Common PAI}}             & \multicolumn{1}{l|}{\multirow{3}{*}{print}}          & \multicolumn{1}{l|}{low}       & \multicolumn{1}{l|}{dpi $\leq$ 600}                       \\ \cline{3-4} 
\multicolumn{1}{|l|}{}                                        & \multicolumn{1}{l|}{}                                & \multicolumn{1}{l|}{medium}    & \multicolumn{1}{l|}{600 $\textless$ dpi $\leq$ 1000}        \\ \cline{3-4} 
\multicolumn{1}{|l|}{}                                        & \multicolumn{1}{l|}{}                                & \multicolumn{1}{l|}{high}      & \multicolumn{1}{l|}{dpi $\textgreater$ 1000}              \\ \cline{2-4} 
\multicolumn{1}{|l|}{}                                        & \multicolumn{1}{l|}{\multirow{3}{*}{replay}}         & \multicolumn{1}{l|}{low}       & \multicolumn{1}{l|}{res $\leq$ 480 pix}                      \\ \cline{3-4} 
\multicolumn{1}{|l|}{}                                        & \multicolumn{1}{l|}{}                                & \multicolumn{1}{l|}{medium}    & \multicolumn{1}{l|}{480 $\textless$ res $\textless$ 1080 pix} \\ \cline{3-4} 
\multicolumn{1}{|l|}{}                                        & \multicolumn{1}{l|}{}                                & \multicolumn{1}{l|}{high}      & \multicolumn{1}{l|}{res $\geq$ 1080 pix}                     \\ \cline{2-4} 
\multicolumn{1}{|l|}{}                                        & \multicolumn{1}{l|}{\multirow{3}{*}{mask}}           & \multicolumn{1}{l|}{paper}     & \multicolumn{1}{l|}{paper masks}                          \\ \cline{3-4} 
\multicolumn{1}{|l|}{}                                        & \multicolumn{1}{l|}{}                                & \multicolumn{1}{l|}{rigid}     & \multicolumn{1}{l|}{non-flexible plaster-like}            \\ \cline{3-4} 
\multicolumn{1}{|l|}{}                                        & \multicolumn{1}{l|}{}                                & \multicolumn{1}{l|}{silicone}  & \multicolumn{1}{l|}{silicone masks}                      \\ \hline \hline
\multicolumn{1}{|l|}{\multirow{6}{*}{Common Capture Devices}} & \multicolumn{1}{l|}{\multirow{2}{*}{webcam}}         & \multicolumn{1}{l|}{low}       & \multicolumn{1}{l|}{SD res}                               \\ \cline{3-4} 
\multicolumn{1}{|l|}{}                                        & \multicolumn{1}{l|}{}                                & \multicolumn{1}{l|}{high}      & \multicolumn{1}{l|}{HD res}                               \\ \cline{2-4} 
\multicolumn{1}{|l|}{}                                        & \multicolumn{1}{l|}{\multirow{2}{*}{mobile/tablet}}  & \multicolumn{1}{l|}{low}       & \multicolumn{1}{l|}{SD res}                               \\ \cline{3-4} 
\multicolumn{1}{|l|}{}                                        & \multicolumn{1}{l|}{}                                & \multicolumn{1}{l|}{high}      & \multicolumn{1}{l|}{HD res}                               \\ \cline{2-4} 
\multicolumn{1}{|l|}{}                                        & \multicolumn{1}{l|}{\multirow{2}{*}{digital camera}} & \multicolumn{1}{l|}{low}       & \multicolumn{1}{l|}{SD res}                               \\ \cline{3-4} 
\multicolumn{1}{|l|}{}                                        & \multicolumn{1}{l|}{}                                & \multicolumn{1}{l|}{high}      & \multicolumn{1}{l|}{HD res}                               \\ \hline \hline
\multicolumn{1}{|l|}{\multirow{3}{*}{Common Face Resolution}} & \multicolumn{1}{l|}{small face}                      & \multicolumn{1}{c|}{-}         & \multicolumn{1}{l|}{IOD $\leq$ 120 pix}                     \\ \cline{2-4} 
\multicolumn{1}{|l|}{}                                        & \multicolumn{1}{l|}{medium face}                     & \multicolumn{1}{c|}{-}         & \multicolumn{1}{l|}{120 $<$ IOD $\leq$ 240 pix}             \\ \cline{2-4} 
\multicolumn{1}{|l|}{}                                        & \multicolumn{1}{l|}{large face}                        & \multicolumn{1}{c|}{-}         & \multicolumn{1}{l|}{IOD $>$ 240 pix}                        \\ \hline \hline
\multicolumn{1}{|l|}{\multirow{3}{*}{Common Lighting}}       & \multicolumn{1}{l|}{controlled}                      & \multicolumn{1}{c|}{-}         & \multicolumn{1}{c|}{-}                                    \\ \cline{2-4} 
\multicolumn{1}{|l|}{}                                        & \multicolumn{1}{l|}{adverse}                         & \multicolumn{1}{c|}{-}         & \multicolumn{1}{c|}{-}                                    \\ \cline{2-4} 
\multicolumn{1}{|l|}{}                                        & \multicolumn{1}{l|}{no info}                         & \multicolumn{1}{c|}{-}         & \multicolumn{1}{c|}{-}                                    \\ \hline
\end{tabular}
\caption{Inter-Dataset common categorization.}
\label{table:categorization}
\end{table}

\begin{enumerate}[leftmargin=3mm]%[I.]
    \setlength{\itemsep}{-0.2\baselineskip}
    \item \textit{Common PAI}. So far, every existing dataset have in common that the PAI used in each attack is precisely labeled. In this work, we also propose an additional sub-categorization in three subtypes based on both quality and material criteria. Print attacks are categorized as: 1) low quality, if the attacks were performed with an image printed on a printer with less than $600 dpi$ (dots per inch), 2) high, if the printer exceeds $1000 dpi$, and 3) medium for the remaining range. The resolution of the screen is used to sub-categorize the Replay attacks as follows: 1) high, if the  attacks are performed replaying a video or a photo on a screen with a resolution higher than $1080 pix$, 2) low, if it is performed on screens with lower resolution than $480 pix$, and 3) medium, as in the previous case, for the remaining range. In the case of mask-type attacks, the categorization is more subjective, so it has been divided on the basis of the material used in each of the masks as: 1) paper, 2) non-flexible plaster-like and 3) silicone. 
    \item \textit{Common capture devices}. Production ready systems that are currently working in the real world have been developed for specific hardware, being able to adjust the parameters to a specific scenario. However, all state-of-the-art systems suffer a dramatic performance loss when they are evaluated in inter-dataset protocols with no reasoning about the parameters involved. GRAD-GPAD allows, in much more detail, an analysis of the influence of these parameters, e.g. the capture devices. Three different types have been added: 1) webcams, 2) mobile/tablet and 3) digital cameras. Additionally, they can be divided by the resolution of their sensors as: high-definition (HD) or standard-definition (SD).
    \item \textit{Common lighting}. Features extracted from color and texture information (crucial in most of current face-PAD methods) are highly influenced by the varying lighting conditions. However, their influence in face-PAD generalization has not been studied properly. In the proposed categorization, we have collected all the information related to lighting. We have found 3 novel divisions: 1) controlled, 2) adverse and 3) no-info. Videos with no information about the lighting have been categorized as no info, waiting for a future categorization.
    \item \textit{Common Face Resolution}: Texture-based methods suffer from different face resolutions. Thus, we propose a inner division in three types: 1) small, 2) medium and 3) large faces. The applied criterion has been extracted from the \textit{ISO/IEC 29794-5:2010} \footnote{https://www.iso.org/standard/50912.html} recommendations, applying thresholds to the Interocular Distance (IOD). Eyes landmarks have been extracted with \cite{mtcnn_zhang_2016} for all the frames available on the aggregated dataset, and the categorization has been made for each video, based on an average of the inter eyes distances of the frames.
\end{enumerate}

\subsection{Protocols}

The GRAD-GPAD framework allows, using configuration files, a simple filtering of data to create our own protocols. We can determine for each of the subsets of the aggregated dataset (Train, Dev and Test) which dataset and subsets are used in the experiment, and filter depending on the common categories selected.

Furthermore, the framework also offers some protocols that have been already proposed in the literature to benchmark the anti-spoofing algorithms. The default protocols are the following:
\begin{enumerate}[leftmargin=3mm]
  \setlength{\itemsep}{-0.4\baselineskip}
  \item \textbf{Grandtest}: a protocol that evaluates face-PAD algorithms without any filter. All sets include all the previous categories.
  \item \textbf{Cross-Dataset}: this is the most common protocol used to assess generalization of face-PAD algorithms. It is based on training on one or several datasets and testing in others.
  \item \textbf{One-PAI}: a protocol that evaluates face-PAD algorithms filtering only by PAI. This protocol is useful to evaluate systems that are focused on the detection of only one kind of PAI.
  \item \textbf{Unseen Attacks (Cross-PAI)}: a protocol that evaluates the performance under unseen PAI. In the training and dev stage a PAI is excluded, which is then used for testing.
  \item \textbf{Unseen Capture Devices}: protocols that evaluate how a face-PAD algorithm works under capture devices that were excluded on training and dev stages.
\end{enumerate}

Despite these protocols already appeared in the literature, the proposed categorization and the aggregated dataset allow a better representation of the anti-spoofing scenario (more examples of different devices, attacks, conditions, etc.). That way, these protocols can extract finer information from the methods under evaluation. Additionally, we propose in this work two novel protocols that evaluate parameters of great relevance in the generalization to real environments.

\begin{enumerate}[leftmargin=3mm]
    \setlength{\itemsep}{-0.25\baselineskip}
	\item \textbf{Cross-FaceResolution}. Building on top of some papers \cite{boulkenafet2018generalization} that mention the influence of the distance between camera and subject, we propose a protocol to focus the analysis on the role of resolution of the region of the face in the task of detecting fake attempts. Two variants are possible for this protocol. The first one, \textit{Cross-FaceResolution-LF-Test}, uses small and medium faces (determined by the resolution) for training, while using high resolution faces (LF - Large Faces) for testing. The second, \textit{Cross-FaceResolution-SF-Test}, represents the opposed setup: it uses large and medium faces samples on training and small faces (SF) for testing.
	\item \textbf{Cross-Conditions}. The underlying idea is to assess the performance of the anti-spoofing algorithms under adversarial conditions. Two variants are proposed: \textit{Cross-Conditions-Test-Adverse}, where we evaluate the performance of the system when training on optimal conditions (high quality capture devices, low and medium quality PAIs, paper masks and both controlled and no info lighting conditions) and  testing on adverse conditions (low quality capture devices, high quality PAIs, silicon and non-flexible plaster-like mask, and adverse lighting conditions); and \textit{Cross-Conditions-Test-Optimal}, where we do the opposite.
\end{enumerate}

%----------------------------------------------------------
\section{Experiments}
\label{sec:experiments}

In this section we present the results of the experiments for the proposed benchmark. The GRAD-GPAD framework is used to extract the features, using two well-known face-PAD methods. We then train the models and filter the categories in order to evaluate several protocols for the proposed aggregated dataset.

\subsection{Face-PAD methods}

 Two recent and popular approaches have been selected to guide the explanation of the proposed protocols, which will also serve as a baseline for the aggregated dataset: 1) the Color-based face-PAD proposed in \cite{color_based_facepad_boulkenafet_2015}, and 2) the Quality-Based method proposed in \cite{anomaly_idiap_2018} based on a concatenation of quality measures (IQM) introduced in \cite{iqa_galbally_2015} and \cite{db_msu_mfsd_2015}. The code for these algorithms is publicly available in the GRAD-PAD framework based on the reproducible material\footnote{https://github.com/zboulkenafet}\footnote{https://gitlab.idiap.ch/bob/bob.pad.face/} shared by the authors. %The GRAD-PAD framework and the interfaces to use the mentioned approaches, are also publicly available.%\footnote{https://github.com/Gradiant/bob.paper.icb2019.gradgpad}.

%\subsection{Preprocessing}

Since we want to compare systems using the same conditions, we fix the video preprocessing and the classification stage for both algorithms. Following the recommendations given in \cite{challenges_face_pad_gradiant_2019}, the experiments have been designed trying to simulate a real scenario and considering usability as a key aspect of the systems.

% Actual deployment constraints, such as frame rate ($FR$) and time of acquisition ($T_a$), have been taken into account. After a visual inspection of existing databases, we realize that most of them have a warm-up period where no action is performed. Thus, we fix the preprocessor of each face-PAD algorithm to use $T_a = 2000 ms$ of the video, but we shift the starting time to $1000ms$ before and after the mid-point of the sequence to avoid the warm-up period. For the sake of usability and the time required to run the experiments, we subsample the image sequences to simulate a capture device with 15 frames-per-second. The preprocessor also extracts faces using a Python implementation of the method proposed in \cite{mtcnn_zhang_2016}. 

For face detection we use the method proposed in \cite{mtcnn_zhang_2016}. %Frames with faces larger than 70 pixels are cropped based on the face location and rescaled to 64 x 64 pixels, otherwise the frame is discarded. 
Faces are cropped and rescaled to 64 x 64 pixels. This cropped image is the input for the method under evaluation. In the case of the Quality-Based algorithm, we obtain for every image a 139-length feature vector from the concatenation of the quality measurements proposed in \cite{iqa_galbally_2015, db_msu_mfsd_2015}. On the other hand, the Color-Based face-PAD computes for each image a vector of a much larger dimensionality (19998-length feature vector) by concatenating texture features (LBP-based) extracted in two color spaces (YCrCb and HSV). Then, for each access video, a feature vector is obtained as the average of the extracted features in each frame. Finally, an SVM classifier with an RBF kernel ($\gamma$ = $1/num  features$) is trained for both systems.

\subsection{Classical Evaluation}

First of all, the face-PAD methods are evaluated with two protocols well known to the community (\textit{Grandtest} and \textit{Cross-Dataset}). The aim of the experiments is two-fold. First, to test the selected algorithms on the aggregated dataset and to provide a baseline for further comparisons. Second, to take advantage of the capabilities of the GRAD-GPAD framework to perform an in-depth analysis of the two approaches. Initially, we evaluate the algorithms with the \textit{Grandtest} protocol.% using the proposed aggregated dataset. %containing the following datasets (ten in total): CASIA-FASD, REPLAY-ATTACK, 3DMAD, MSU-MFSD, REPLAY-MOBILE, HKBU, OULU-NPU, ROSE-YOUTU, SIW, CSMAD.

Now, from the results in Table \ref{results:grantest} we can easily observe that the Color-based approach performs better in general. Besides, the large difference, in both approaches, between HTER and ACER, suggests the influence of some PAIs.

\begin{table}[htbp]
\centering
\resizebox{0.77\textwidth}{!}{
	\begin{minipage}{\textwidth}
		\begin{tabular}{c|c|c|c|c|}
		\cline{2-5}
		\multicolumn{1}{l|}{}               & \multicolumn{1}{l|}{HTER (\%)} & \multicolumn{1}{l|}{ACER (\%)} & \multicolumn{1}{l|}{APCER (\%)} & \multicolumn{1}{l|}{BPCER (\%)} \\ \hline \hline
		\multicolumn{1}{|c|}{Quality-Based} & 17.03                          & 25.25                          & 34.09                           & 16.41                           \\ \hline
		\multicolumn{1}{|c|}{Color-Based}   & 6.33                           & 10.22                          & 13.86                           & 6.58                            \\ \hline
		\end{tabular}
	\end{minipage}
}
\caption{Results for \textit{Grandtest} protocol}
\label{results:grantest}
\end{table}

In order to understand the source of the error regarding the PAI, we have used the One-PAI protocol. As this protocol has only one type of attack for testing, the reported metric is HTER (equivalent to ACER for the single case). The Quality-based face-PAD obtains better results for replay (HTER = 10.22\%) and mask attacks (HTER = 10.08\%), compared with  print attacks (HTER = 14.53\%). The color-based approach, on the contrary, has a slightly more stable behavior with a better performance for mask attacks (HTER = 2.90\%), while print and replay attacks get HTER values of 4.41\% and 3.92\% respectively.

Once we have clarified the strengths and weaknesses of the two approaches in the aggregated dataset, it is time to check their generalization capabilities. Table \ref{results:crossdataset} represents the results for leave-one-out \textit{Cross-Dataset} protocol, where the named dataset is used for testing the generalization of the models and the nine remaining datasets are used for training and tuning the algorithms.

\begin{table}[htbp]
      \centering
      \tiny
      \begin{subtable}{.5\textwidth}
             \centering
             \begin{tabular}{|c|c|c|c|c|}
                \hline
                Test          & HTER (\%)  & ACER (\%)  & APCER (\%) & BPCER (\%) \\ \hline \hline
                CASIA-FASD    & 41.57      & 48.98      & 81.11      & 16.85      \\ \hline
                REPLAY-ATTACK & 27.61      & 34.06      & 33.96      & 34.17      \\ \hline
                3DMAD         & 29.00      & 29.00      & 0.00       & 58.00      \\ \hline
                MSU-MFSD      & 31.11      & 46.66      & 46.66      & 46.66      \\ \hline
                REPLAY-MOBILE & 26.89      & 28.19      & 34.37      & 22.02      \\ \hline
                HKBU          & 45.00      & 45.00      & 90.0       & 0.00       \\ \hline
                OULU-NPU      & 34.68      & 41.11      & 75.27      & 6.94       \\ \hline
                ROSE-YOUTU    & 37.88      & 45.81      & 42.40      & 49.22      \\ \hline
                SIW           & 31.97      & 48.40      & 53.07      & 43.74      \\ \hline
                CSMAD         & 40.51      & 40.51      & 10.20      & 70.83      \\ \hline
              \end{tabular}
      \caption{Results using the Quality-Based face-PAD}
      \label{results:crossdataset:quality}
      \end{subtable}
      \begin{subtable}{.5\textwidth}
          \centering
          \begin{tabular}{|c|c|c|c|c|}
              \hline
              Test          & HTER (\%)  & ACER (\%)  & APCER (\%) & BPCER (\%) \\ \hline \hline
              CASIA-FASD    & 15.45      & 16.75      & 17.78      & 15.73      \\ \hline
              REPLAY-ATTACK & 25.11      & 33.35      & 31.25      & 35.44      \\ \hline 
              3DMAD         & 0.00       & 0.00       & 0.00       & 0.00       \\ \hline
              MSU-MFSD      & 17.78      & 35.00      & 56.66      & 13.33      \\ \hline
              REPLAY-MOBILE & 18.30      & 22.99      & 23.96      & 22.02      \\ \hline
              HKBU          & 0.00       & 0.00       & 0.00       & 0.00       \\ \hline
              OULU-NPU      & 34.27      & 37.78      & 72.22      & 3.33       \\ \hline
              ROSE-YOUTU    & 27.42      & 34.78      & 25.25      & 44.32      \\ \hline
              SIW           & 9.90       & 22.06      & 30.43      & 13.69      \\ \hline
              CSMAD         & 40.05      & 40.05      & 55.10      & 25.00      \\ \hline
              \end{tabular}

      \caption{Results using the Color-Based face-PAD}
      \label{results:crossdataset:color}
      \end{subtable}

\caption{Results for \textit{Cross-Dataset} protocol}
\label{results:crossdataset}
\end{table}

To extract realistic conclusions from the results of the cross-dataset protocol, we have to take into account the global parameters of each of the datasets (see Table \ref{table:datasets}). From the experiments we observe that the Color-based method generalizes consistently better in all the datasets considered in the cross-dataset protocol. It even gets remarkable results in 3DMAD and HKBU databases. The common element between these two databases is that all attacks are performed using rigid masks. It seems that the Color-based method (texture features over different color spaces) achieves fair performance for this type of attacks. However, none of the methods generalizes properly in scenarios with different PAIs, as in all of them the ACER is above 22\%. In this classical protocol there are many factors that are involved in an uncontrolled way and, without additional information, it is very complicated to draw fair conclusions. It is therefore necessary to carry out evaluations focusing on fewer parameters.

\subsection{Proposed Evaluation: the new protocols}

The proposed benchmark has three protocols that allow us to draw conclusions from another perspective: the \textit{Cross-Devices} protocol evaluates performance on unseen capture devices; the \textit{Cross-FaceResolution} protocol measures the effects of variations in face resolution; and the \textit{Cross-Conditions} protocol evaluates the influence of adversarial operating conditions.

Table \ref{results:crossdevice} shows the results of Quality-Based and Color-Based approaches using the Cross-Device protocol, leaving out one type of capture device (digital camera, webcam or mobile/tablet as named in the table) for training and development stages and using it on testing. Such results indicate that the most challenging scenario for the two systems evaluated is \textit{Cross-Device-Webcam-Test}. This result seems to be closely related to the quality of the capture device, as webcams have the worst image quality. We can conclude that there is a drop in the performance of both face-PAD systems as the image quality degrades. In addition, it can be appreciated how, with this type of protocols, the Quality-based method suffers much more than the Color-based one. %This behavior seems consistent as the Quality-based method is much more sensitive to the image quality degradation. 
%The Quality-based method even obtains worse results than chance. However, the Color-based method presents quite a good performance for \textit{Cross-Device-MobileTablet-Test} protocol with an ACER of 12.30\%.

\begin{table}[htbp]
\resizebox{0.60\textwidth}{!}{
	\begin{minipage}{\textwidth}
		\begin{tabular}{c|c|c|c|c|c|}
		\cline{2-6}
		                                                     & \begin{tabular}[c]{@{}c@{}}\textit{Cross-Device}\\Protocol\end{tabular}                                            & HTER (\%) & ACER (\%) & APCER (\%) & BPCER (\%) \\ \hline \hline
		\multicolumn{1}{|c|}{\multirow{3}{*}{Quality-Based}} & DigitalCamera-Test                     & 24.85     & 52.27     & 86.67      & 17.89      \\ \cline{2-6} 
		\multicolumn{1}{|c|}{}                               & Webcam-Test                            & 28.55     & 53.57     & 29.52      & 47.62      \\ \cline{2-6} 
		\multicolumn{1}{|c|}{}                               & MobileTablet-Test					  & 21.11     & 25.76     & 29.33      & 22.19      \\ \hline
		\multicolumn{1}{|c|}{\multirow{3}{*}{Color-Based}}   & DigitalCamera-Test                     & 7.42      & 16.26     & 26.76      & 5.75       \\ \cline{2-6} 
		\multicolumn{1}{|c|}{}                               & Webcam-Test                            & 12.16     & 31.90     & 48.98      & 14.83      \\ \cline{2-6} 
		\multicolumn{1}{|c|}{}                               & MobileTablet-Test                      & 9.08      & 12.30     & 15.07      & 9.54       \\ \hline
		\end{tabular}
	\end{minipage}
}
\caption{Results for \textit{Cross-Device} protocol}
\label{results:crossdevice}
\end{table}

To compliment the study of capture devices, it is worth to carry out another interesting experiment, taking into account the image resolution. Besides, if we also had the information of the resolution of the detected face available, we could extract very interesting properties to figure out the optimal distance for realistic scenarios. Results using the \textit{Cross-FaceResolution} protocol are shown in Table \ref{results:crossfaceresolution}. We can observe that the Color-based method on \textit{Cross-FaceResoultion-LF-Test} is the only experiment within this protocol that seems to achieve good generalization capabilities. In view of these results, it seems that there is a better capacity for generalization when we move towards higher resolution scenarios, having been trained with lower resolution, than in the opposite case (for the Color-based face-PAD). %Note that, as explained in the description of the methods, the cropped faces are resized to the same size of 64 x 64, so filtering is done using the original resolution of the input image. 
 
\begin{table}[htbp]
\resizebox{0.58\textwidth}{!}{
	\begin{minipage}{\textwidth}
		\begin{tabular}{c|c|c|c|c|c|}
		\cline{2-6}
		                                                     & \begin{tabular}[c]{@{}c@{}}\textit{Cross-FaceResolution}\\Protocol\end{tabular}                     & HTER (\%) & ACER (\%) & APCER (\%) & BPCER (\%) \\ \hline \hline
		\multicolumn{1}{|c|}{\multirow{2}{*}{Quality-Based}} & LF-Test & 24.48     & 51.86     & 86.21      & 17.52      \\ \cline{2-6} 
		\multicolumn{1}{|c|}{}                               & SF-Test & 29.98     & 48.79     & 50.00      & 47.58      \\ \hline
		\multicolumn{1}{|c|}{\multirow{2}{*}{Color-Based}}   & LF-Test & 8.33      & 15.81     & 27.50      & 4.12       \\ \cline{2-6} 
		\multicolumn{1}{|c|}{}                               & SF-Test & 25.47     & 29.62     & 12.20      & 47.04      \\ \hline
		\end{tabular}
	\end{minipage}
}
\caption{Results for \textit{Cross-FaceResolution} protocol}
\label{results:crossfaceresolution}
\end{table}

If we analyze the results reported in both Table \ref{results:crossdevice} and Table \ref{results:crossfaceresolution}, we can see that the most favorable case for the color-based method occurs when we train using webcams and digital cameras where the resolution of the detected faces are medium and low, and testing on mobile or tablets with a large face resolution. The results for the described combination are an ACER of 51.77\% for the Quality-based method, while it is 39.07\% in the case of Color-based. A smaller number of samples for training (more filtering) and a high dimensionality of the features can be the cause of this drop in performance.

Finally, the results of the \textit{Cross-Condition} protocol are presented in Table \ref{results:crossconditions}. It seems that none of the baselines are able to generalize in this protocol. Results indicate that the working point shift is very large, reaching APCER values close to or above 90\%. This novel protocol is therefore postulated as one of the greatest challenges for the aggregated dataset presented.

\begin{table}[htbp]
\resizebox{0.61\textwidth}{!}{
	\begin{minipage}{\textwidth}
		\begin{tabular}{c|c|c|c|c|c|}
		\cline{2-6}
		                                                     & \begin{tabular}[c]{@{}c@{}}\textit{Cross-Conditions}\\Protocol\end{tabular}                      & HTER (\%) & ACER (\%) & APCER (\%) & BPCER (\%) \\ \hline \hline
		\multicolumn{1}{|c|}{\multirow{2}{*}{Quality-Based}} & Test-Adverse & 36.62     & 40.48     & 72.50      & 8.46       \\ \cline{2-6} 
		\multicolumn{1}{|c|}{}                               & Test-Optimal & 45.50     & 66.06     & 96.67      & 35.46      \\ \hline
		\multicolumn{1}{|c|}{\multirow{2}{*}{Color-Based}}   & Test-Adverse & 41.13     & 45.43     & 86.25      & 4.61       \\ \cline{2-6} 
		\multicolumn{1}{|c|}{}                               & Test-Optimal & 34.37     & 55.12     & 93.33      & 16.91      \\ \hline
		\end{tabular}
	\end{minipage}
}
\caption{Results for \textit{Cross-Conditions} protocol}
\label{results:crossconditions}
\end{table}

Through the experiments carried out, we have been able to analyze two methods that achieve state-of-the-art results when evaluated in individual databases and observe that, however, they do not manage to generalize adequately in conditions closer to real scenarios. But the information provided by the framework goes further, allowing us to systematically analyze the performance of such systems in the face of other variations such as changes in capture devices, image resolution or extreme mismatches between display conditions. This allows for a fairer and more consistent evaluation and comparison of face-PAD systems.

%----------------------------------------------------------
\section{Conclusions}
\label{sec:conclusions}

In this work we have proposed a framework, GRAD-GPAD, for systematic evaluation of the generalization properties of face-PAD methods. The GRAD-GPAD framework allows the aggregation of heterogeneous datasets, the inclusion of new feature-extraction algorithms and new evaluation protocols.
We have studied the generalization capabilities of two well-known face-PAD methods using a large aggregated dataset comprising ten publicly available datasets, and several protocols, including two new ones proposed in this work. The GRAD-GPAD framework allows to learn some deficiencies of these algorithms that could eventually drive to more robust and generalized feature representations for face-PAD. In short, this paper highlights the importance of data and the necessity of fair evaluation methodologies to improve the generalization of existing face-PAD methods. 

% In general, face anti-spoofing has been treated as a two-class classification problem (real and attacks). Since spoofing attempts in the real world can be so diverse and the capture devices are constantly evolving, we should be aware that each of the available datasets represent a very restricted view of the problem. In this paper we propose a new paradigm within the face-PAD problem that deals with the evaluation of the generalization properties, that we call Generalized Presentation Attack Detection (GPAD), which is considered as the most critical point in face anti-spoofing. We propose a benchmark, based on aggregated datasets, that unifies existing datasets combining their information adequately. Moreover, we propose new methodologies and protocols that let us draw conclusions from a new perspective. 
% We introduce the GRAD-GPAD framework that contains useful tools for the creation of new protocols, such as those presented in this work. The GRAD-GPAD framework allows researchers to learn the deficiencies of their algorithms, giving a step forward to robust and generalized feature representations for face-PAD.

% This paper highlights the importance of data and the necessity of fair evaluation methodologies to improve the generalization of existing face-PAD methods. The proposed Aggregated-Dataset is created with the intention of increasing the size and variability of the data and to provide a shared categorization among existing datasets. 

\paragraph{Acknowledgments} We thank our colleagues of the Biometrics Team at Gradiant for their valuable contributions.

{\small
\bibliographystyle{ieee}
\bibliography{egbib}

\begin{thebibliography}{10}\itemsep=-1pt

\bibitem{motion_anjos_2011}
A.~Anjos and S.~Marcel.
\newblock Counter-measures to photo attacks in face recognition: A public
  database and a baseline.
\newblock In {\em International Joint Conference on Biometrics}, 2011.

\bibitem{db_csmad_2018}
S.~Bhattacharjee, A.~Mohammadi, and S.~Marcel.
\newblock {Spoofing Deep Face Recognition With Custom Silicone Masks}.
\newblock In {\em {Biometrics: Theory, Applications, and Systems}}, 2018.

\bibitem{color_based_facepad_boulkenafet_2015}
Z.~Boulkenafet, J.~Komulainen, and A.~Hadid.
\newblock Face spoofing detection using colour texture analysis.
\newblock {\em IEEE Transactions on Information Forensics and Security (TIFS)},
  2016.

\bibitem{boulkenafet2018generalization}
Z.~Boulkenafet, J.~Komulainen, and A.~Hadid.
\newblock On the generalization of color texture-based face anti-spoofing.
\newblock {\em Image and Vision Computing}, 2018.

\bibitem{db_oulu_npu_2017}
Z.~Boulkenafet, J.~Komulainen, L.~Li, X.~Feng, and A.~Hadid.
\newblock {OULU-NPU}: A mobile face presentation attack database with
  real-world variations.
\newblock 2017.

\bibitem{db_replay_attack_2012}
I.~Chingovska, A.~Anjos, and S.~Marcel.
\newblock On the effectiveness of local binary patterns in face anti-spoofing.
\newblock In {\em BIOSIG}, 2012.

\bibitem{db_replay_mobile_2016}
A.~Costa-Pazo, S.~Bhattacharjee, E.~Vazquez-Fernandez, and S.~Marcel.
\newblock The replay-mobile face presentation-attack database.
\newblock In {\em BioSIG}, 2016.

\bibitem{challenges_face_pad_gradiant_2019}
A.~Costa-Pazo, E.~Vazquez-Fernandez, J.~L. Alba-Castro, and
  D.~González-Jiménez.
\newblock Challenges of face presentation attack detection in real scenarios,
  2019.

\bibitem{can_face_antispoofing_defreitas2013}
T.~de~Freitas~Pereira, A.~Anjos, J.~M.~D. Martino, and S.~Marcel.
\newblock Can face anti-spoofing countermeasures work in a real world scenario?
\newblock In {\em ICB 2013}, 2013.

\bibitem{db_3dmad_2013}
N.~Erdogmus and S.~Marcel.
\newblock Spoofing in 2d face recognition with 3d masks and anti-spoofing with
  kinect.
\newblock 2013.

\bibitem{iqa_galbally_2015}
J.~Galbally, S.~Marcel, and J.~Fierrez.
\newblock Image quality assessment for fake biometric detection: Application to
  iris, fingerprint, and face recognition.
\newblock {\em IEEE Transactions on Image Processing}, 2014.

\bibitem{jackson2017vrn}
A.~S. Jackson, A.~Bulat, V.~Argyriou, and G.~Tzimiropoulos.
\newblock Large pose 3d face reconstruction from a single image via direct
  volumetric cnn regression.
\newblock {\em International Conference on Computer Vision}, 2017.

\bibitem{db_siw_2018}
A.~Jourabloo*, Y.~Liu*, and X.~Liu.
\newblock Face de-spoofing: Anti-spoofing via noise modeling.
\newblock In {\em Proc. European Conference on Computer Vision}, Munich,
  Germany, 2018.

\bibitem{kollreider2007real}
K.~Kollreider, H.~Fronthaler, M.~I. Faraj, and J.~Bigun.
\newblock Real-time face detection and motion analysis with application in
  “liveness” assessment.
\newblock {\em IEEE TIFS}, 2007.

\bibitem{Korshunova_2017_ICCV}
I.~Korshunova, W.~Shi, J.~Dambre, and L.~Theis.
\newblock Fast face-swap using convolutional neural networks.
\newblock In {\em ICCV}, 2017.

\bibitem{li2018learning}
H.~Li, P.~He, S.~Wang, A.~Rocha, X.~Jiang, and A.~C. Kot.
\newblock Learning generalized deep feature representation for face
  anti-spoofing.
\newblock {\em IEEE TIFS}, 2018.

\bibitem{db_rose_youtu_2018}
H.~Li, W.~Li, H.~Cao, S.~Wang, F.~Huang, and A.~C. Kot.
\newblock Unsupervised domain adaptation for face anti-spoofing.
\newblock {\em IEEE TIFS}, 2018.

\bibitem{db_hkbu_2016}
S.~Liu, P.~C. Yuen, S.~Zhang, and G.~Zhao.
\newblock 3d mask face anti-spoofing with remote photoplethysmography.
\newblock In B.~Leibe, J.~Matas, N.~Sebe, and M.~Welling, editors, {\em
  Computer Vision -- ECCV}. Springer International Publishing, 2016.

\bibitem{maatta2011face}
J.~M{\"a}{\"a}tt{\"a}, A.~Hadid, and M.~Pietik{\"a}inen.
\newblock Face spoofing detection from single images using micro-texture
  analysis.
\newblock In {\em IJCB 2011}. IEEE, 2011.

\bibitem{db_smad_2017}
I.~Manjani, S.~Tariyal, M.~Vatsa, R.~Singh, and A.~Majumdar.
\newblock Detecting silicone mask-based presentation attack via deep dictionary
  learning.
\newblock {\em IEEE TIFS}, 2017.

\bibitem{anomaly_idiap_2018}
O.~Nikisins, A.~Mohammadi, A.~Anjos, and S.~Marcel.
\newblock On effectiveness of anomaly detection approaches against unseen
  presentation attacks in face anti-spoofing.
\newblock In {\em ICB}, 2018.

\bibitem{db_uvad_2015}
A.~Pinto, W.~R. Schwartz, H.~Pedrini, and A.~d.~R.~Rocha.
\newblock Using visual rhythms for detecting video-based facial spoof attacks.
\newblock {\em IEEE TIFS}, 2015.

\bibitem{light_field_based_sepas_2018}
A.~Sepas-Moghaddam, F.~Pereira, and P.~L. Correia.
\newblock Light field-based face presentation attack detection: Reviewing,
  benchmarking and one step further.
\newblock {\em IEEE TIFS}, 2018.

\bibitem{tirunagari2015detection}
S.~Tirunagari, N.~Poh, D.~Windridge, A.~Iorliam, N.~Suki, and A.~T. Ho.
\newblock Detection of face spoofing using visual dynamics.
\newblock {\em IEEE TIFS}, 2015.

\bibitem{wang2018exploiting}
Z.~Wang, C.~Zhao, Y.~Qin, Q.~Zhou, and Z.~Lei.
\newblock Exploiting temporal and depth information for multi-frame face
  anti-spoofing.
\newblock {\em arXiv preprint arXiv:1811.05118}, 2018.

\bibitem{db_msu_mfsd_2015}
D.~Wen, H.~Han, and A.~Jain.
\newblock { Face Spoof Detection with Image Distortion Analysis}.
\newblock {\em IEEE TIFS}, 2015.

\bibitem{face-upad}
F.~Xiong and W.~Abdalmageed.
\newblock Unknown presentation attack detection with face rgb images.
\newblock In {\em BTAS}, 2018.

\bibitem{mtcnn_zhang_2016}
K.~Zhang, Z.~Zhang, Z.~Li, and Y.~Qiao.
\newblock Joint face detection and alignment using multi-task cascaded
  convolutional networks.
\newblock {\em CoRR}, 2016.

\bibitem{zhang2018casia}
S.~Zhang, X.~Wang, A.~Liu, C.~Zhao, J.~Wan, S.~Escalera, H.~Shi, Z.~Wang, and
  S.~Z. Li.
\newblock Casia-surf: A dataset and benchmark for large-scale multi-modal face
  anti-spoofing.
\newblock {\em arXiv preprint arXiv:1812.00408}, 2018.

\bibitem{db_casia_surf_2018}
S.~Zhang, X.~Wang, C.~Zhao, J.~Wan, S.~Escalera, H.~Shi, Z.~Wang, and S.~Z.Li.
\newblock A dataset and benchmark for large-scale multi-modal face
  anti-spoofing.
\newblock In {\em In Conference on Computer Vision and Pattern Recognition},
  2019.

\bibitem{db_casia_fasd_2012}
Z.~Zhang, J.~Yan, S.~Liu, Z.~Lei, D.~Yi, and S.~Z. Li.
\newblock A face antispoofing database with diverse attacks.
\newblock In {\em ICB 2012}, 2012.

\end{thebibliography}
}

\end{document}